\documentclass{article}

\PassOptionsToPackage{numbers, compress}{natbib}

\usepackage[preprint]{nips_2018}

\usepackage[utf8]{inputenc} 
\usepackage[T1]{fontenc}    
\usepackage{hyperref}       
\usepackage{url}            
\usepackage{booktabs}       
\usepackage{amsfonts}       
\usepackage{nicefrac}       
\usepackage{microtype}      
\usepackage{graphicx}

\usepackage{caption}
\usepackage{subcaption}
\usepackage{wrapfig,lipsum,booktabs}

\title{Why every GBDT speed benchmark is wrong}

\author{
  Anna Veronika  Dorogush \And Vasily Ershov \And Dmitriy Kruchinin 
}

\begin{document}

\maketitle

\begin{abstract}
    This article provides a comprehensive study of different ways to make speed benchmarks of gradient boosted decision trees (GBDT) algorithm. We show main problems of several straight forward ways to make benchmarks, explain, why a speed benchmark is challenging task and provide a set of reasonable requirements for a benchmark to be fair and useful.
\end{abstract}

\section{Introduction}

Gradient boosting is a powerful machine-learning technique that achieves state-of-the-art results in a variety of practical tasks. Over the number of years, it remains the primary method for learning problems with heterogeneous features, noisy data, and complex dependencies: web search, recommendation systems, weather forecasting, and many others~\cite{caruana2006empirical,roe2005boosted,wu2010adapting,zhang2015gradient}. Most popular implementations of gradient boosting use decision trees as base predictors. The three most known highly-efficient open-source GBDT libraries are XGBoost~\cite{chen2016xgboost}, LightGBM\cite{lightgbm}, CatBoost\cite{dorogush2017fighting}.

One of the first questions a potential user asks is how fast is the algorithm and what level of quality could be reached at the fixed problem. One can find a huge number of benchmarks for speed and quality from users on Kaggle, research papers, and GBDT library authors. As developers of CatBoost\footnote{Project site: \url{https://catboost.ai}},  and we are interested in the community making benchmarks for their own tasks on a fair equal basis. Thus we are searching for a single routine everyone will agree on.

Exploring existing benchmarks and conducting our own ones, we faced many problems and found out that most of the openly available comparisons have different mistakes.

Some of them are technical, for example, measuring disk access time instead of learning time, or using small data as an example of big data problems, or making a comparison on a single dataset with an unusual bottleneck, which will not be reproduced on other datasets.

Here we will address a more fundamental problem: why, even after solving the technical issues, most GBDT speed benchmarks would not be trustworthy. In the next two sections, we will describe several suggested approaches that could be used to estimate GBDT speed performance, and provide problems one can face with applying them.

\section{Why speed benchmark is challenging task}

The main problem of every benchmark is to design an experiment that will not give an advantage to any algorithm. At the same time, the experiment should be done in a meaningful way: it should report a useful result.
For example, it's desirable to account for the achieved quality of each algorithm: otherwise, one could just sample random trees on each iteration and be faster than others. 
Here we introduce the list of the most popular ideas of creating  ``meaningful'' speed benchmarks for GBDT:

\begin{enumerate}
    \item Freezing the parameter set and the number of iterations for each algorithm.
    \item Measuring time to train until optimal or some fixed quality is reached.
\end{enumerate}

Training time depends on per iteration time and on a number of iterations. Per iteration time depends on data characteristics, such as a number of columns and rows, sparsity level and number of categorical features. And a number of iterations to convergence to a fixed quality is highly dependant on the problem we are trying to solve, even if we have two datasets of similar sparsity and size.

\paragraph{Freezing the parameter set and the number of iterations for each algorithm}

All GBDT libraries have a similar set of hyperparameters: number of iterations, maximum depth of a tree, learning rate, objects subsampling rate, and others. One of the simplest ways to measure performance is to freeze the same set of parameters for each library. The result of the experiment is the time to train $n$ trees with depth\footnote{Or number of leaves $2^k$ in case of LightGBM} $k$ and some metric value that represents achieved quality of the model. But the resulting quality values are not meaningful and using them in benchmarks can be misleading.

\begin{figure}
\centering
\begin{subfigure}{.45\textwidth}
 \centering
  \includegraphics[width=1.1\linewidth]{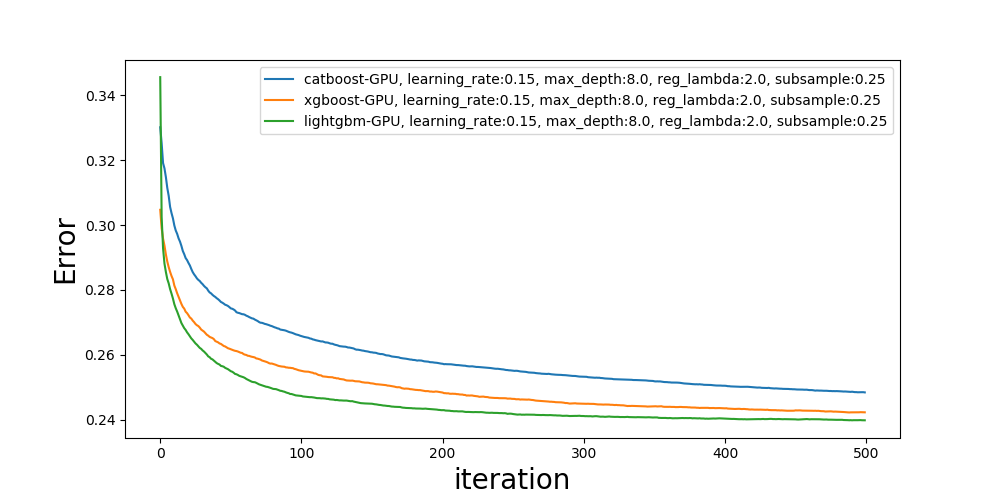}
  \caption{ 500 iterations}
  \label{fig:higgs500}
\end{subfigure}%
\begin{subfigure}{.45\textwidth}
 \centering
  \includegraphics[width=1.1\linewidth]{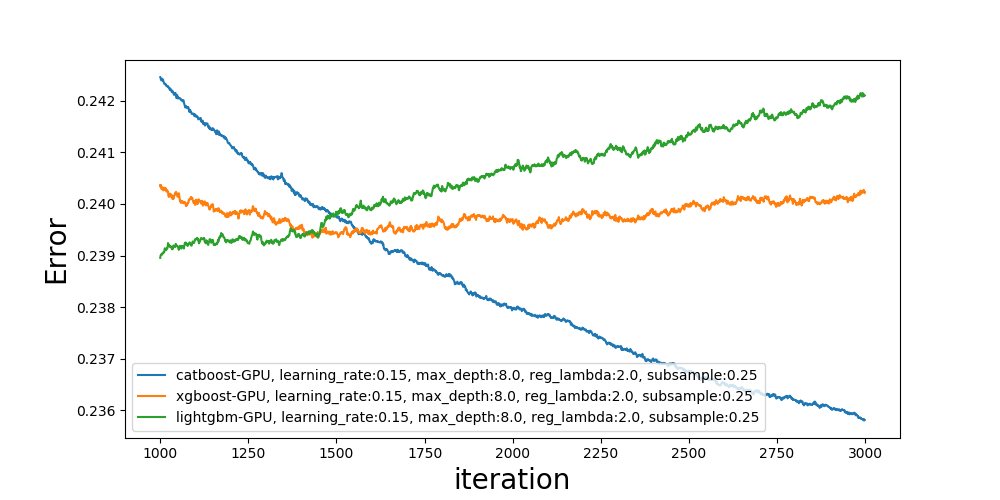}
  \caption{3000 iterations}
  \label{fig:higgs3000}
\end{subfigure}%
\caption{Higgs: XGBoost vs LigthGBM vs CatBoost with same params}
\end{figure}

Different libraries use a different type of growing policies for decision trees, and, as a result, their parameters are not comparable. For example, we've trained all three libraries on Higgs dataset with similar parameters\footnote{Max tree depth~-- 8, learning rate~-- 0.15, L2 regularization parameter~-- 2.0, object subsampling~-- 0.25}. Figures~\ref{fig:higgs500},\ref{fig:higgs3000} show metric values for the first 500 iterations and for iterations from 1000 to 3000. As you can see, the winner in terms of quality has changed twice after 1500 iteration. This shows that the quality benchmark done in this fashion is misleading. We need to do more, than simply using the same parameters, if we want to make a meaningful comparison.

Nevertheless we still get some good insights about GBDT library speed: boosting is an iterative algorithm and time to train a fixed number of trees is a robust metric. We expect to see similar training time of the algorithm if we will change dataset to a similar one\footnote{by similar we mean with approximately equal number of features, their sparsity levels and number of samples}. To proof this we've measured time we need to run 500 iterations for each library on 3 different dense datasets: row-subsampled Higgs, column-subsampled Epsilon and sklearn synthetic. All three datasets have 28 features and 500000 rows. Table~\ref{tab:fit_500} shows, that time we need to fit 500 iterations of depth 6 trees is similar across datasets for each library. 
\begin{table}[h!]
\vspace{-5pt}
\scriptsize
    \centering
    \begin{tabular}{|c|c|c|c|}
    \hline
    \textbf{Dataset} & \textbf{CatBoost} & \textbf{XGBoost} & \textbf{LightGBM} \\\hline
    Synthetic & 5.22 & 10.94 & 27.66\\\hline
    Epsilon col-sampled  & 5.48 & 9.95  & 39.61\\\hline
    Higgs row-sampled    & 5.24 & 11.43 & 27.58\\\hline
    \end{tabular}
 \vspace{3pt}
    \caption{Time to fit 500 iterations on dense datasets with the same shape (28 features, 500k samples)}
    \label{tab:fit_500}
      \vspace*{-15pt}
\end{table}

Thus, such benchmarks could be useful if we will skip quality metrics. For example, in real-world problems, we often know the amount of time we could spend fitting the model. And with information from this benchmark one could select a strategy for parameter tuning for a given dataset.

\paragraph{Measuring time to train until the optimal quality is reached}

\begin{figure}[ht]
    \vspace*{-15pt}
    \begin{subfigure}{0.32\textwidth}
        \centering
        \includegraphics[width=\linewidth]{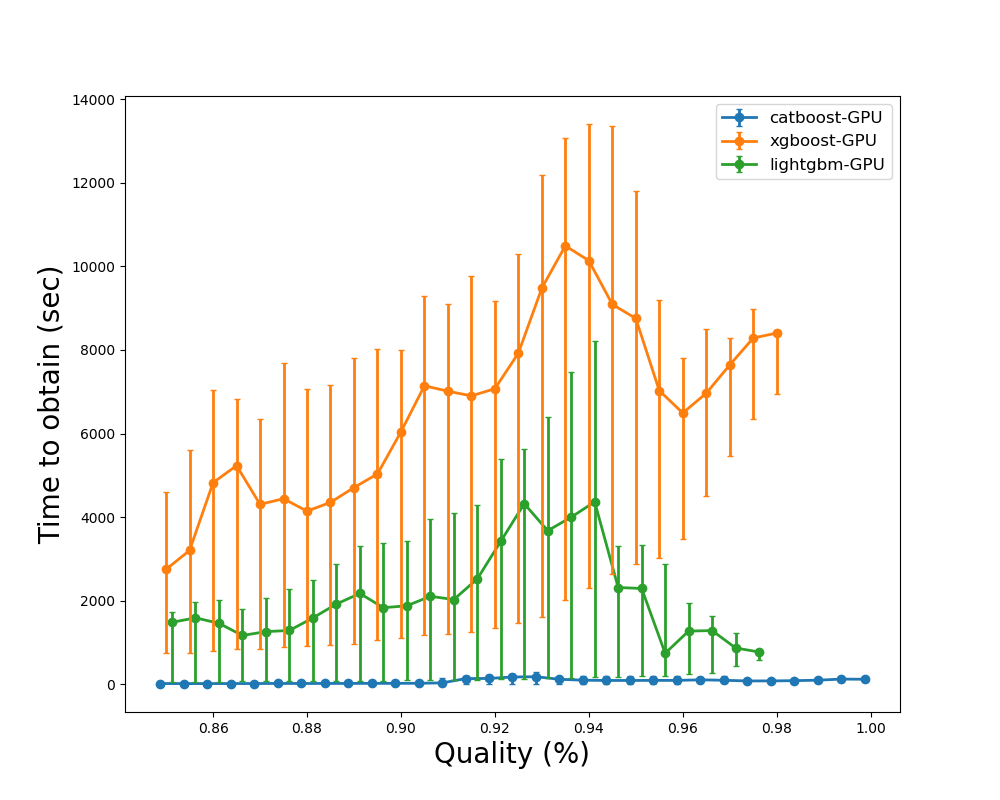}
        \caption{Time to reach quality: Epsilon}
        \label{fig:epsilon_min_time}
    \end{subfigure}%
    \begin{subfigure}{0.32\textwidth}
        \centering
        \includegraphics[width=\linewidth]{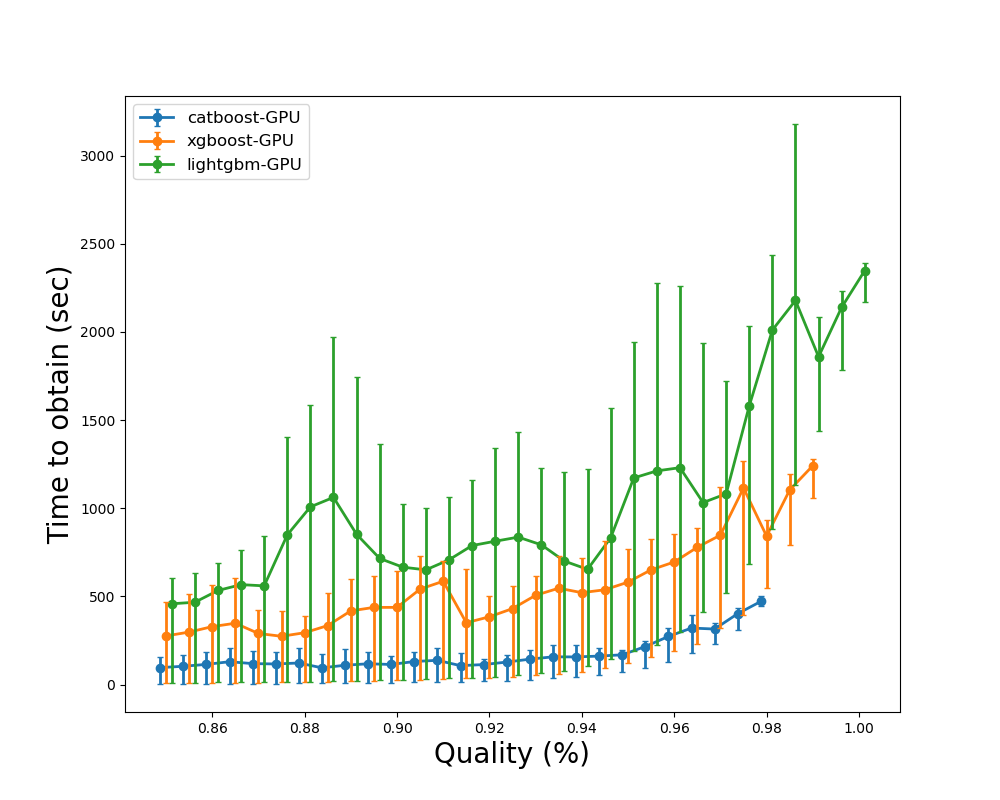}
        \caption{Time to reach quality: Higgs}
        \label{fig:higgs_min_time}
    \end{subfigure}%
    \begin{subfigure}{0.32\textwidth} 
    \centering
      \includegraphics[width=\linewidth]{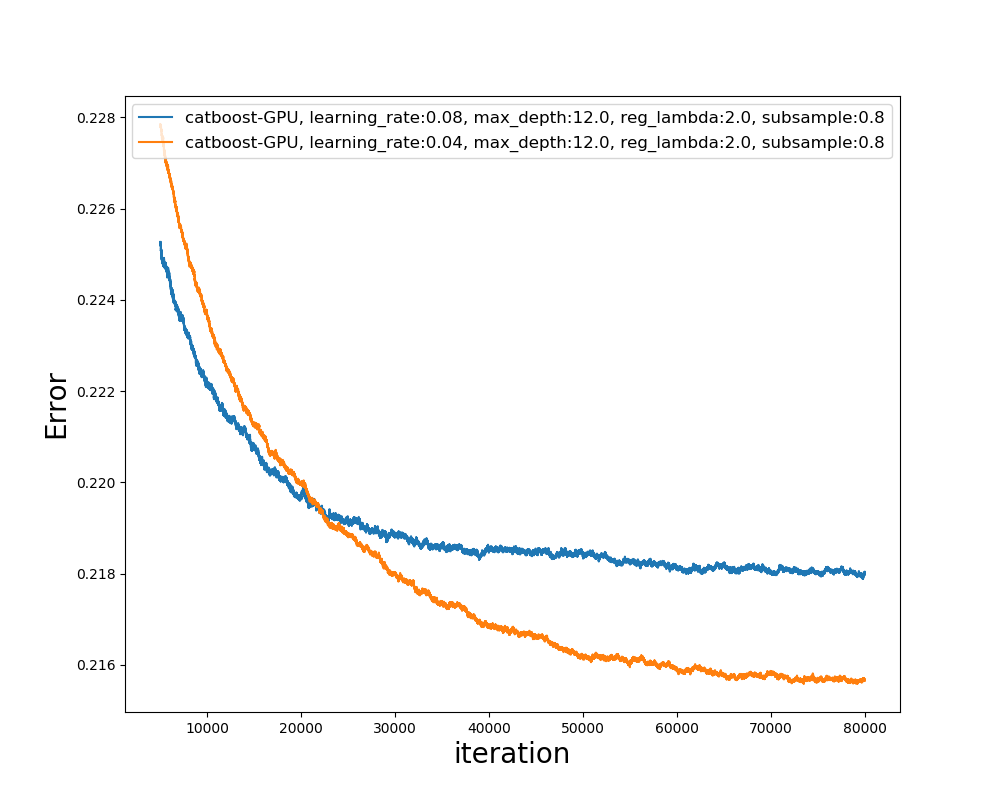}
      \caption{Huge iterations: Higgs}
      \label{fig:higgs}
    \end{subfigure}%
\end{figure}

The next idea is to measure the training time of the algorithm until its best possible quality is reached. This time might include the parameter tuning process, or ignore it.

When training GBDT, decreasing learning rate often provides an improvement in quality. When dealing with a large dataset, we could train models with tens of thousands of estimators, and this will provide quality improvement. Let's look on Higgs dataset example. There are three key parameters that mostly affect quality: tree depth, ensemble size, and learning rate. In this case, we can train models with 10k iterations and then decrease learning rate by a factor of 2, increase the total number of iterations and achieve better quality, as you could see from learning curves on figure~\ref{fig:higgs}. 

This means, that we can't measure the time that algorithm needs to achieve an optimal quality: in many cases, it's just not possible to make a descent grid-search or bayesian optimization for hyperparameters in a reasonable amount of time, so we won't know, what is the optimal quality for our dataset.

\paragraph{Measuring time to train until some fixed quality is reached}

In many cases changes in the tenth decimal place of a metric may not interest us (is $0.1\%$ of a quality metric worth it to train model 3-4 times slower?), small fluctuations could be invisible in practice and therefore irrelevant. Thus the comparison ``learn until fixed quality'' can either be done by an expert with domain knowledge, and be valid only for this particular case, or it should be done with different quality marks.

Each library provides different ways to improve learning speed in exchange for slight drop in overall quality, these ways could significantly differ from one problem to another. For example, let's look on figures~\ref{fig:epsilon_min_time}, \ref{fig:higgs_min_time}. 
We run all 3 libraries on two datasets on grids listed in table~\ref{tab:params_grid}\footnote{We've used GPU version of all libraries, experiments were done on: NVIDIA GTX1080TI for Epsilon; NVIDIA Titan V for Higgs}. 
We've calculated the best quality we were able to achieve, and fixed a grid of relative percents of this value. For each level of quality we've filtered all experiments from grid search that reach it and computed median (circle point), minimum and maximum time when algorithm achieve that score. As you can see, we could achieve the same quality with different parameters in significantly different times.

This is another example of a misleading benchmark. We've used the same parameter grid and the fixed amount of iterations, but we've already discussed, that parameters are not comparable with each other. To make this benchmark fair we need to solve hyper parameter optimization problem:
find an optimal set of parameters that will allow achieving this fixed quality in the minimum amount of time. But such task is even harder than searching for the best possible parameters. We come back to the same problem of misleading quality numbers as before. And like the previous one, this benchmark will also highly depend on a concrete problem: as you can see, plots from Higgs and Epsilon experiments are not similar at all.



It is worth noting that there are practical tasks that don't have a fixed dataset, so we can gather as much data as library could handle in a fixed amount of time or/and on a fixed hardware. So, if the library works worse on the same dataset but is two times faster, we could make a bigger dataset and reach the same quality mark. 

\begin{wraptable}{R}{6cm}
  \vspace*{-10pt}
\tiny
    \centering
    \begin{tabular}{|l|l|l|}
    \hline
    Parameter & Higgs & Epsilon\\\hline
    \texttt{max\_depth} & [6, 8, 10] & [6, 8, 10]\\\hline
    \texttt{learn\_rate}  & [.01, .03, .07, .15, .3] & [.03, .07, .15]\\\hline
    \texttt{subsample} & [.25, .5, .75, 1] & [.5, .75, 1]\\\hline
    \texttt{reg\_lambda} & [2, 4, 8, 16] & [1]\\\hline
    \end{tabular}
     \vspace*{-2pt}
    \caption{Parameters grid for Higgs and Epsilon, we've run all algorithm for 5k iterations and choose best}
    \label{tab:params_grid}
    \vspace*{-13pt}
\end{wraptable}

\section{Conclusion}

As you can see, benchmarking speed for GBDT is a hard task. Each library works better in different setups and on different datasets. The most useful speed benchmark should account for all the problems we've mentioned. We believe that good benchmark should use different datasets to study pros and cons of each library. For each dataset this benchmark should not just report a single number, but provide comprehensive study of each boosting implementation and show what  speed vs quality  trade-offs could be made. This benchmark should not decide for our users what they need, but give them a chance to make their own decision, based on their limitations in terms of training time or prediction time, and on how much do they care about quality tuning. 

The problem is, it's impossible to create such benchmark in a reasonable amount of time. And we often need only rough comparison of libraries performances. Benchmark that freeze parameters to similar ones, is a good example of such rough comparison. It's not fair, but if we'll stop reporting quality numbers, such benchmark could become useful.

\small
\bibliography{boosting}
 
\bibliographystyle{abbrv}
\end{document}